\documentclass[10pt,twocolumn,letterpaper]{article}

\usepackage[pagenumbers]{cvpr} 

\usepackage{graphicx}
\usepackage{amsmath}
\usepackage{multirow}
\usepackage{amssymb}
\usepackage{booktabs}
\usepackage{float}
\usepackage{newtxmath}
\usepackage{color, xcolor}
\usepackage{soul}

\usepackage{xcolor}

\usepackage[pagebackref,breaklinks,colorlinks]{hyperref}

\usepackage[capitalize]{cleveref}
\crefname{section}{Sec.}{Secs.}
\Crefname{section}{Section}{Sections}
\Crefname{table}{Table}{Tables}
\crefname{table}{Tab.}{Tabs.}

\def\cvprPaperID{6088} 
\def\confName{CVPR}
\def\confYear{2022}

\begin{document}

\title{DGECN: A Depth-Guided Edge Convolutional Network for \\
End-to-End 6D Pose
Estimation}

\renewcommand{\thefootnote}{\fnsymbol{footnote}}
\author{Tuo Cao\textsuperscript{1},
Fei Luo\textsuperscript{1}\footnotemark[1],
Yanping Fu\textsuperscript{2},
Wenxiao Zhang\textsuperscript{1},
Shengjie Zheng\textsuperscript{1},
and Chunxia Xiao\textsuperscript{1}\footnotemark[1]\\
\textsuperscript{1}{School of Computer Science, Wuhan University, Wuhan, Hubei, China}\\
\textsuperscript{2}{ School of Computer Science and Technology,
Anhui University, Hefei, Anhui, China}\\
{\tt\small ypfu@ahu.edu.cn, wenxxiao.zhang@gmail.com, zsj\_mdk@163.com, \{maplect,luofei,cxxiao\}@whu.edu.cn}\\
{\small \url{http://graphvision.whu.edu.cn/}}
}
\maketitle

\begin{abstract}
   Monocular 6D pose estimation is a fundamental task in computer vision. Existing works often adopt a two-stage pipeline by establishing correspondences and utilizing a RANSAC algorithm to calculate 6 degrees-of-freedom (6DoF) pose. Recent works try to integrate differentiable RANSAC algorithms to achieve an end-to-end 6D pose estimation. However, most of them hardly consider the geometric features in 3D space, and ignore the topology cues when performing differentiable RANSAC algorithms. To this end, we proposed a  \textbf{D}epth-\textbf{G}uided \textbf{E}dge \textbf{C}onvolutional \textbf{N}etwork (DGECN) for 6D pose estimation task. We have made efforts from the following three aspects: 1) We take advantages of estimated depth information to guide both the correspondences-extraction process and the cascaded differentiable RANSAC algorithm with geometric information. 2) We leverage the uncertainty of the estimated depth map to improve accuracy and robustness of the output 6D pose. 3) We propose a differentiable Perspective-n-Point(PnP) algorithm via edge convolution to explore the topology relations between 2D-3D correspondences. Experiments demonstrate that our proposed network outperforms current works on both effectiveness and efficiency.
\end{abstract}

\section{Introduction}
Object pose estimation is a task of calculating the $6$ degrees of freedom (DoF) pose of a rigid object, including its location and orientation in an image. It is widely used in the three-dimensional registration of AR~\cite{xiang2017posecnn,peng2019pvnet,andrew2001multiple}, robotic vision~\cite{rad2017bb8,park2019pix2pose} and 3D reconstruction~\cite{fu2020joint,fu2021seamless}. Due to the presence of noises and other influential factors, such as the occlusion, noisy background, and illumination variations, accurately estimating the 6DoF poses of the objects in the RGB image is still a challenging problem. 
\begin{figure}[t]
\centering
   \includegraphics[width=1.0\linewidth]{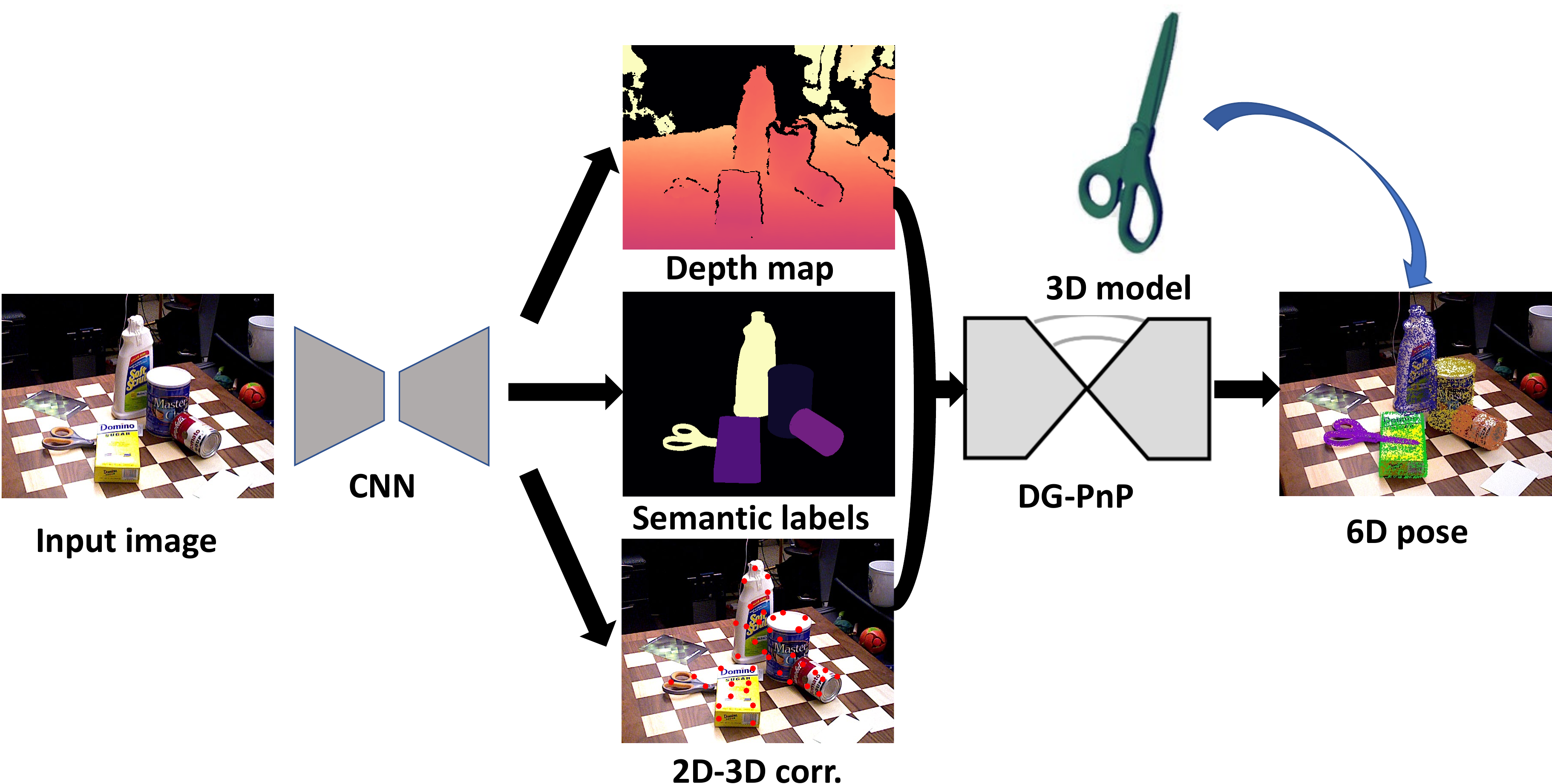}
      \caption{{\bf Pipeline of DGECN}. With an input RGB image, we propose a novel DGECN to simultaneously predict segmentation and depth maps. After established 2D-3D correspondences, we replace the RANSAC/PnP with a learnable DG-PnP to regress 6D pose.}
\label{fig:1}
\end{figure}

Current object pose estimation methods can be divided into two types: 1) the object poses are estimated using a single RGB image~\cite{xiang2017posecnn,hu2019segmentation,peng2019pvnet,park2019pix2pose,rad2017bb8} or 2) an RGB image accompanying a depth image~\cite{he2020pvn3d,wada2020morefusion,wang2019densefusion}. For both RGB based and RGB-D based methods, the keypoints-based works are dominant in this field. On the other hand, methods based on direct regression are usually inferior to keypoints-based methods. The keypoints-based methods usually consist of two stages: firstly it predicts the 2D location of the keypoints of the 3D model on RGB images via a modern neural network. And then calculate the 6D pose parameters with the RANSAC-based Perspective-n-Point (PnP) method from 2D-3D correspondences. Although many representative works~\cite{hodan2020epos,li2019cdpn,manhardt2019explaining,rozumnyi2020sub,shao2020pfrl,song2020hybridpose} have proven the validity of the two-stage pipeline, there are still many limitations in it. Firstly, few methods can directly output the 6D pose parameters. Most of the existing methods still use a variant of the RANSAC-based PnP algorithm to estimate the pose parameters. Secondly, RANSAC-based PnP can be very time-cost when the 2D-3D correspondences are dense. Thirdly, the network in most two-stage works cannot directly output 6D pose, so their loss functions cannot optimize our expected pose estimation. Finally, the two-stage estimation may lead to significant accumulative error, which gradually increases among the two connected steps.

Recently, some works try to integrate a differentiable RANSAC algorithm into the pipeline, so the network can be trained end-to-end. Brachmann \etal~\cite{Brachmann_2017_CVPR} proposed a differentiable PnP method. Hu \etal~\cite{hu2020single} leveraged PointNet~\cite{qi2017pointnet} to approximate PnP for sparse correspondences. But these works either require a cumbersome training process or do not consider the geometry clues. Wang \etal~\cite{Wang_2021_CVPR} made an end-to-end framework by replacing RANSAC-based PnP with Patch-PnP, this method works well, but it relies on the Dense Correspondences Map and Surface Region Attention Map in their network. It can hardly directly learning 6D pose from 2D-3D correspondences. 

To this end, we propose Depth-Guided Edge Convolutional Network (DGECN), jointly handling the correspondences extraction and the 6D pose estimation. Our network leverages a depth guided network to establish 2D-3D correspondences and learn the 6D pose from the correspondences by a novel Dynamic Graph PnP (DG-PnP). 
On one hand, depth information allows us to make full use of the geometric constraint of rigid objects. On the other hand, we fully revisit the properties of correspondence set and find it can better handle complex textures by constructing a graph structure.
Our end-to-end pipeline is shown in Fig.~\ref{fig:1}.

Experimental results on LM-O~\cite{brachmann2014learning} and YCB-V~\cite{calli2015ycb,xiang2017posecnn} demonstrate our network is comparable even superior to the state-of-the-art methods in terms of accuracy and efficiency.

Our contributions in this work can be summarized as follows:
\begin{itemize}
\item We propose a Depth-guided network to directly learn the 6D pose from a monocular image without additional information required. Furthermore, we propose a Depth Refinement Network (DRN) to polish the quality of the estimated depth map.
\end{itemize}
\begin{itemize}
\item We explore the properties of 2D correspondence sets and discover that 6D pose parameters can be learned better from the 2D keypoint distributions by constructing a graph. We further propose a simple but effective Dynamic Graph PnP (DG-PnP) to directly learn 6D pose from 2D-3D correspondences. 
\end{itemize}

\section{Related work}

\noindent \textbf{Direct Methods.} These methods usually directly estimate the 6D pose in a single shot. Some early works leverage template matching techniques. However, they do not perform satisfactorily under occlusion. With the advance of deep learning, some works regress the pose parameters via a network. Xiang.\etal~\cite{xiang2017posecnn} first introduced CNN into this field, they employed a network based on GoogleNet~\cite{szegedy2015going} to directly learn the 6D camera pose. This problem is still challenging due to the variety of objects as well as the complexity of a scene caused by clutter and occlusions between objects. To address this flaw, PoseCNN~\cite{xiang2017posecnn} estimated the 3D translation of an object by localizing its center in the image and predicting its distance from the camera. However, this problem is still difficult due to the non-closed property to addition of rotation matrix. Some works~\cite{zhou2019continuity} utilized the $\mathbb{SO}(3)/\mathbb{SE}(3)$ to make the rotation space differentiable. 

\noindent \textbf{Correspondence-based Methods.} The methods based on 2D-3D correspondence detection have gradually become the mainstream in object pose estimation. 
PVNet~\cite{peng2019pvnet} and Seg-Driven~\cite{hu2019segmentation} conducted segmentation coupled with voting for each correspondence to make the estimation more robust. EPOS~\cite{hodan2020epos} made use of surface fragments accounting for ambiguities in pose. Pix2Pose~\cite{park2019pix2pose} used a network based on GAN to predict the 3D coordinates of each object pixel without textured models. Oberweger \etal~\cite{oberweger2018making} output pixel-wise heatmaps of keypoints to address the issue of occlusion. 
 Recent years, a few works aim to avoid the time-consuming RANSAC-based PnP in keypoint-based pipeline. DSAC~\cite{Brachmann_2017_CVPR} presented two alternative ways to make RANSAC differentiable by soft argmax and probabilistic selection and applied it to the problem of camera localization. Single-Stage~\cite{hu2020single} employed a PointNet-like architecture to learn the 6D pose from 2D-3D correspondences. However, this method can only deal with the sparse correspondences. To avoid this, GDR-Net~\cite{Wang_2021_CVPR} let the network predict the surface regions as additional ambiguity-aware supervision and used them within their Patch-PnP framework. SO-Pose~\cite{Di_2021_ICCV} focused on the occluded part to encode the geometric features of the object more completely and accurately.

\noindent\textbf{Graph Convolution Network (GCN).} Due to the higher representation power of graph structure, GCN has demonstrated superior performance in several tasks, including image caption~\cite{dong2021dual}, text to image and human pose estimation~\cite{cai2019exploiting}. In 3D computer vision, Wald \etal~\cite{Wald_2020_CVPR} proposed the first learning method that generated a semantic scene graph from a 3D point cloud. DGCNN~\cite{wang2019dynamic} used a GCN-based network for point cloud feature extraction. Superglue~\cite{sarlin2020superglue} leveraged GCN to match two sets of local features by jointly searching correspondences and rejecting non-matchable points.

\begin{figure*}
\centering
\setlength{\abovecaptionskip}{0pt}
\includegraphics[width=6.9in]{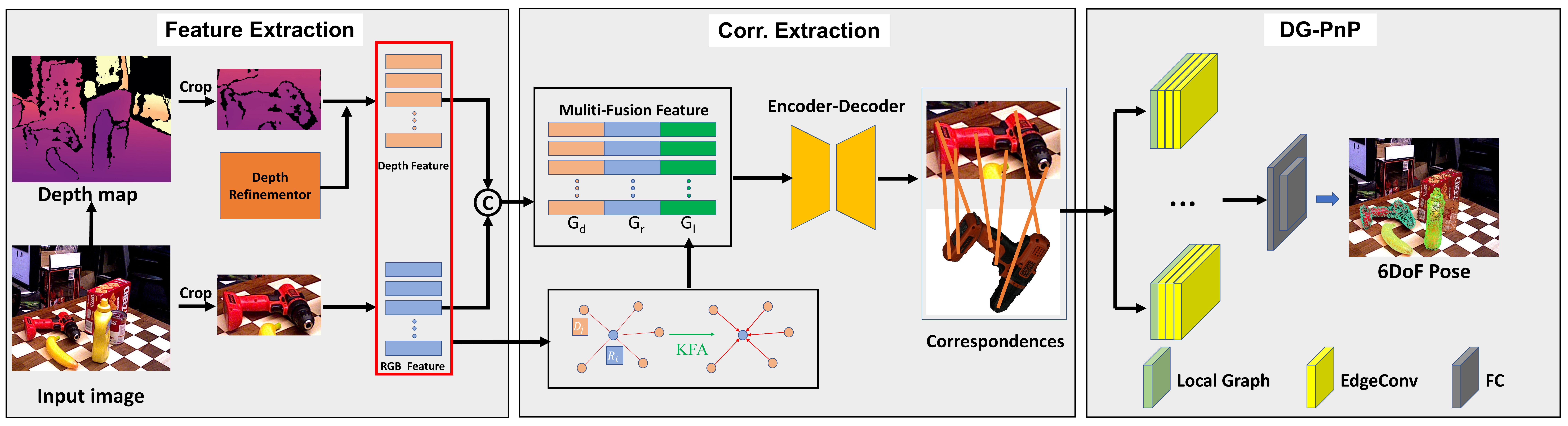}
\caption{{\bf Overview of our architecture.} Our framework consists of three building blocks: 1) a feature extraction network for depth and RGB features fusion, 2) a 2D-3D correspondences extraction network based on a deep voting-based network, and 3) a learnable PnP network named DG-PnP for 6D pose object estimation. KFA means K-NN Feature Aggregation, $G_r$, $G_d$ and $G_l$ are RGB feature, depth feature and local feature, respectively.}
\label{fig:2}
\end{figure*}

\section{Approach}
\label{sec:3}

In this section, we will describe our depth-guided 6D pose regression network. We first introduce the relevant background. Then, we illustrate our network architecture which can learn the depth to refine the 6D pose. 

\subsection{Problem Formulation}

Given an image, our task is to detect the objects and estimate the 6D pose of them. Here, we denote the image as $I$. Our goal is to estimate the rotation $\mathbf{R}\in \mathbb{SO}(3)$ and translation $\mathbf{t}=(t_x, t_y, t_z)\in \mathbb{R}^{3}$ that can transform the object from its object world coordinate system to the camera world coordinate system. 

Fig~\ref{fig:2} is the overview of our proposed method. We first learn depth information via an unsupervised depth estimation network. Afterwards, like GDR-Net~\cite{Wang_2021_CVPR} and PVNet~\cite{peng2019pvnet}, we locate each object in the image with the method of FCN~\cite{long2015fully}. According to the results of the segmentation, we crop the region of interest on depth map and RGB image, and fed them to a K-NN based feature aggregation (KFA) module to get the local features. Meanwhile, we use ResNet50~\cite{he2016deep} to extract the 2D features of the image. Then, a dense fusion module is used to fuse the appearance features, geometry information and local features. Next, we take the fused feature as input of a 2D-3D correspondences prediction network to establish the 2D-3D correspondences. Finally, we directly regress the associated 6D object pose from the 2D-3D correspondences via our proposed differentiable DG-PnP.

Our framework builds upon keypoint-based methods.
Given an image $I$ and 3D models $M = \{M_i |i=1,...,N\}$, our objective is to recover the unknown rigid transformation $\{\mathbf{R}, \mathbf{t}\}$. For the convenience of display, we assume that there is one target object in the image, we denote it as $O$. As shown in Fig~\ref{fig:3}, our goal is to predict the potential 2D location in $I$ of the corresponding 3D keypoints of the model $M$. 

\subsection{Depth Estimation}
Inspired by recent works~\cite{he2020pvn3d,wang2019densefusion,zhang2019pcan,zhang2020detail} based on RGB-D data and point cloud, we introduce depth information to make 2D-3D correspondences more robust and accurate. However, these methods always need LIDAR or other sensors to get true depth information. Moreover, in a beforehand acquired RGB image, we usually can not obtain true depth information. Therefore, we use a network to predict the depth as an additional feature to supervise the 2D-3D correspondences estimation. With the development of monocular depth estimation, many depth estimation methods~\cite{ref_Mono2,Ramamonjisoa_2021_CVPR,ref_Hints} have emerged. However, these methods are often used to estimate the depth information of large scenes, which is not good to directly estimate the depth map of 6D pose estimation scene. Therefore, in our work, we use uncertainty measurement to refine the estimated depth map. 

\begin{figure}
\centering
   \includegraphics[width=1.0\linewidth]{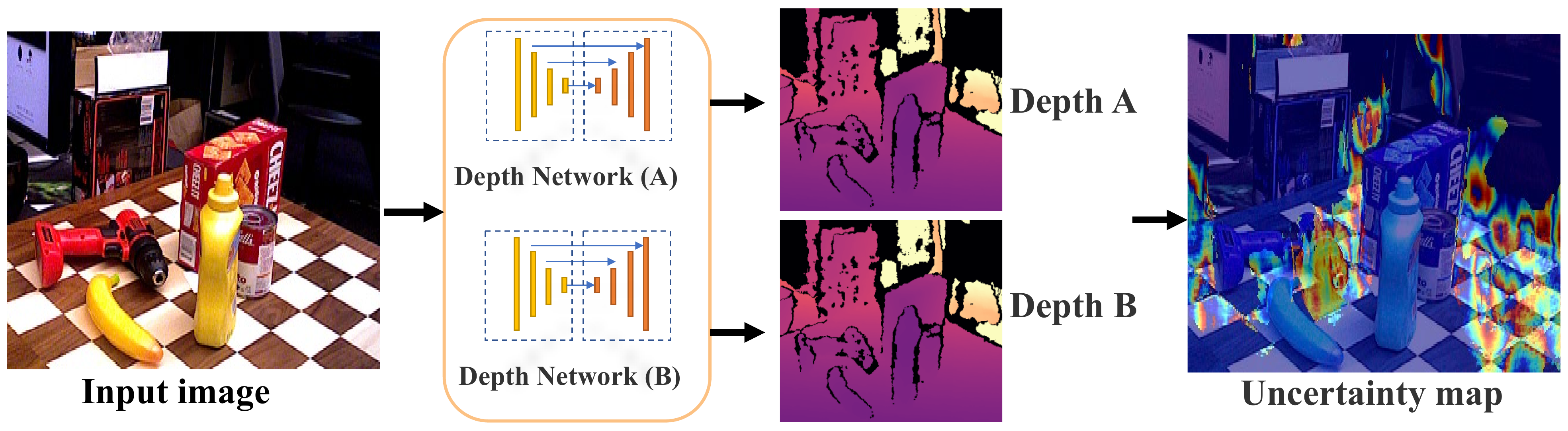}
   \caption{{\bf Depth Uncertainty Measurement}.}
\label{fig:um}
\end{figure}

\subsection{Depth-Guided Edge Convolutional Network}
The overview of our method is shown in Fig.~\ref{fig:2}. The keypoint localization is a voting-based architecture, which does not fully consider depth information. Therefore, we have made efforts in three directions to improve this strategy:
\begin{enumerate}
\item We leverage the uncertainty of estimated depth map on 6D object estimation scenes, we refine the depth map and reduce the influence of noise in the depth estimation process.
\item Before directly feeding RGB into CNN for establishing 2D-3D correspondences, we firstly predict the depth map and propose a K-NN Feature Aggregation (KFA) block to fuse cross-domain features.
\item We propose a learnable DG-PnP to replace the handcrafted RANSAC/PnP in the two-stage 6d pose estimation pipeline.
\end{enumerate}

\noindent\textbf{Depth refinement network (DRN).}
Current monocular depth estimation methods are often applied to large outdoor scenes. Therefore, they are usually trained on large scenes dataset, such as KITTI. However, when we directly use these methods to estimate 6DoF scenes depth, in some areas, the fluctuations may be particularly large. The DRN aims to polish the quality of the depth map. As shown in Fig~\ref{fig:um}, it is composed of two different depth estimation networks, each network output a depth map $D_A$ and $D_B$, respectively. We then calculate the difference between two depth maps, and define the area where the difference is over the threshold as an uncertain area. There are two ways to further handle these uncertain areas, one is directly remove them from the depth feature. The second way is to use their mean to replace the original depth. We choose the first way in this paper. 

\noindent \textbf{Feature extraction.}
This stage has two streams, one for depth estimation and the other for object segmentation. Depth estimation takes a color image as input and performs depth map prediction. Then, for each segmented object, we use the segmented object mask and the depth map to convert it to a 3D point cloud. To deal with multiple objects segmentation, previous works~\cite{peng2019pvnet,xiang2017posecnn,wang2019densefusion,hu2019segmentation} use existing detection or semantic segmentation algorithms. Similarly, we adopt FCN~\cite{long2015fully} to segment the input image. As for 3D feature extraction, some works~\cite{wang2019densefusion,he2020pvn3d} convert the segmented depth pixels into a 3D point cloud, and the utilize 3D feature extractor~\cite{qi2017pointnet,qi2017pointnet++,DBLP:journals/corr/abs-2012-09688} to extract geometric features. Although these methods are proved to be effective, they need to train additional 3D feature network. 
For more sufficient RGB-D fusion, we introduce KFA module. Consider a pixel in RGB image, denoted as $p_i$, and $D_i = \{d_j|j=1...k\}$ is a depth set of the k-nearest neighbors of $p_i$, then we adopt a nonlinear function $F_{p_i}=f(D_i,\theta_i)$ with a learnable parameter $\theta_i$ to aggregate the local feature of $p_i$. As shown in Fig.~\ref{fig:2}, the resulting feature $G=(G_r,G_d,G_l)$.

\begin{figure}[t]
\centering
   \includegraphics[width=1.0\linewidth]{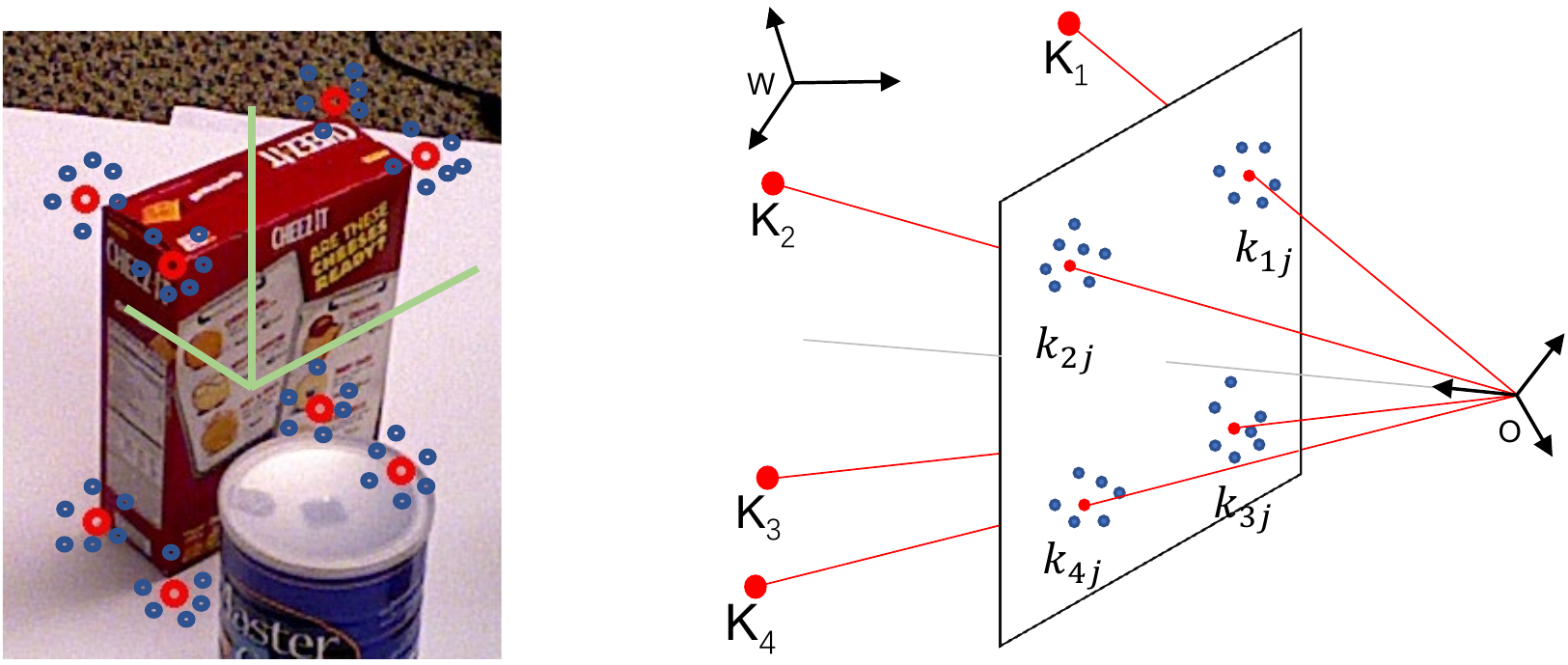}
    \\[-0.2ex]

(a)\hspace{4cm}(b) 
\\[-0.2ex]
   \caption{{\bf 2D-3D correspondences}. \textbf{(a)} Ground truth 2D correspondences (red ones) with their hypotheses (blue ones). \textbf{(b)} 2D correspondences projections on the camera plane. The camera and object coordinate sys- tems are denoted by $O$ and $W$, respectively.}
  
\label{fig:3}
\end{figure}

\noindent \textbf{2D keypoint localization.} The 3D keypoints are selected from the 3D object model as in~\cite{he2020pvn3d,peng2019pvnet}. Some methods~\cite{hu2019segmentation,rad2017bb8} choose the eight corners of the 3D bounding box. However, these points are virtual and 2D correspondences may locate outside the image. For the object closed to the boundary, this may lead to large errors, since the 2D correspondences are not in the image. Therefore, the keypoints should be selected on the object surface. We follow~\cite{peng2019pvnet} and adopt the farthest point sampling (FPS) algorithm to select keypoints on object surface. At the end of this stage, we use a network based on~\cite{hu2019segmentation} for 2D correspondences detection.

\noindent \textbf{Learning 6D pose from 2D-3D correspondences.} As shown in Fig.~\ref{fig:3}, given a set with $n$ 3D keypoints $K=\{K_i|i=1,...,n\}$ and each $K_i$ corresponds to a set of 2D locations $k = \{k_{ij}|j=1,...,m\}$ in image. Our goal is to design a network to learn the rigid transformation $(\mathbf{R}, \mathbf{t})$ from the established 2D-3D correspondences. DSAC~\cite{Brachmann_2017_CVPR} made RANSAC differentiable by soft argmax and probabilistic selection. Single-Stage~\cite{hu2020single} utilized a PointNet-like architecture to address this, however it can only handle sparse correspondences. GDR-Net~\cite{Wang_2021_CVPR} proposed a simple but effective patch-PnP module, where it depends on the dense correspondences maps that predicted by their network. To handle this, we propose a GCN based network to directly regress the 6D pose from the 2D-3D correspondences, which is described as follows
\begin{equation}\label{eq1}
(\mathbf{R}, \mathbf{t})=\mathcal{M}(K,k|\Theta),
\end{equation}
where $\mathcal{M}$ denotes the proposed DG-PnP with parameters $\Theta$.

Hu \etal\cite{hu2020single} used an architecture similar to PointNet~\cite{qi2017pointnet}. However, it only takes the 2D location as individual point and does not take into account the distribution property of 2D correspondences in the image. As mentioned above, we predict depth value of every pixel in the input image, therefore we can make full use of the geometric and location features of 2D correspondences. By revisiting the properties of 2D-3D correspondences, we find that the structure of the 2D correspondences is similar to a graph. As shown in Fig.~\ref{fig:3}, instead of taking individual points as input, we take the 2D correspondence cluster as a graph and feed it into our DG-PnP. 

\begin{figure}[t]
\centering
   \includegraphics[width=1.0\linewidth]{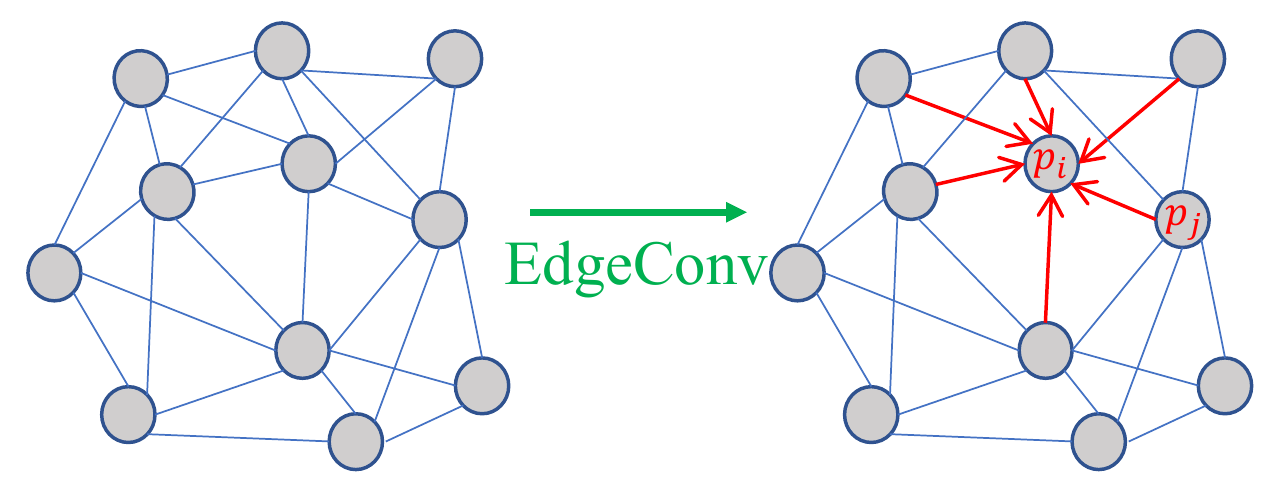}
   \caption{{\bf Local graph and edge convolution}.}
\label{fig:lg}
\end{figure}

\noindent \textbf{Local Graph Construction.} As shown in Fig.~\ref{fig:lg},  $\mathcal{P}=\{p_i|i=1...m\}$ is a 2D correspondences cluster, we construct the local graph via k-nearest neighbor (k-NN) and denote it as $\mathcal{G}=(\mathcal{P},\mathcal{E})$. $\mathcal{P}$ and $\mathcal{E}=p_i\leftrightarrow p_j$ are vertices and edges, respectively. Then, we compute edge features by aggregating all neighborhoods of $p_i$ in $\mathcal{P}$.

\noindent \textbf{Edge-convolution.} Different from graph convolution network (GCN), our edge-convolution is a variant of CNN. Considering a 2D correspondence cluster of $m$ pixels with $X$ dimension features, and denoting it as $f=\{f_i|i=1,...,m\}$, we compute the local graph feature by our graph operation:
\begin{equation}
f^{'}_i = \sum^{m}_{j=1}\lambda_j g_{\theta_i}(f_i,f_j),
\end{equation}
where $\lambda_j$ is a hyperparameter which is determined by the distance between $k_i$ and $k_j$. $g_\theta$ is a nonlinear function with a learnable parameters $\theta$. We adopt an asymmetric edge function proposed in~\cite{wang2019dynamic}: 
\begin{equation}
g_{\theta_i}(f_i,f_j) = \text{RELU}(\alpha_i \cdot(f_i - f_j) + \beta_i \cdot f_i),
\end{equation}
where $\theta_i = (\alpha_i , \beta_i)$ and $\Theta = \{ \theta_i| i=1,...,m\}$ in Eq.~\ref{eq1}. In this paper, we take the 3D coordinates and RGB information of $k_i$ as features $f_i$, and the 3D coordinates can be transformed from depth using camera intrinsic. Therefore, $X=6$ in our network. 
 
\subsection{Loss Function and Pose Estimation}

To train the proposed network, we introduce four loss functions $\mathcal{L}_d$, $\mathcal{L}_s$, $\mathcal{L}_k$, and $\mathcal{L}_p$. The total loss function is defined as 
\begin{equation}
\mathcal{L}=\lambda_1\mathcal{L}_d+\lambda_2\mathcal{L}_s+\lambda_3\mathcal{L}_k+\lambda_4\mathcal{L}_p,
\end{equation}
where $\lambda_{1}$, $\lambda_{2}$, $\lambda_3$ and $\lambda_{4}$ are the weight coefficients.

$\mathcal{L}_d$ is the depth loss, and depth estimation module is built upon MonoDepth2~\cite{ref_Mono2}:
\begin{equation}
\mathcal{L}_d = \mu L_{dp}+\lambda L_{ds},
\end{equation}
where $L_{dp}$ is photometric loss, and $L_{ds}$ is edge-aware smoothness. Due to space limitations, further details can refer to ~\cite{ref_Mono2}.

$\mathcal{L}_s$ is the segmentation loss, which is used to constrain the segmentation task and extract the target object from the image. Here we choose the Focal Loss according to ~\cite{lin2017focal}.

$\mathcal{L}_{k}$ is the keypoint matching loss, which is used to constrain the 2D-3D correspondences. As shown in Fig.~\ref{fig:3}, we seek to predict 2D keypoints location in the image and we define the loss function as:
\begin{equation}
\mathcal{L}_k = \frac{1}{M}\sum_{i=1}^{n}\sum_{j=1}^{m}||kp_{ij} - kp_{i}^*||,
\end{equation}
where $kp^*_{i}$ is the ground truth 2D keypoint location, $n$ is the number of 3D keypoints, $m$ is the number of 2D correspondences of $kp_i$, $M=m\times n$ is the number of total 2D correspondences predicted by our network in the image.

$\mathcal{L}_p$ is the final pose estimation loss, which is used to constrain the final 6DoF pose parameters. Inspired by PoseCNN~\cite{xiang2017posecnn} and DeepIM~\cite{li2018deepim}, we design $\mathcal{L}_p$ as 
\begin{equation}
\mathcal{L}_{p}=\frac{1}{n} \sum_{i=1}^{n} \|\left(\mathbf{R}^* \mathbf{p}_{i}+\mathbf{t}^*\right)-\left(\mathbf{R} \mathbf{p}_{i}+\mathbf{t}\right) \|.
\end{equation}
where $\mathbf{R}^*$ and $\mathbf{t}^*$ are the estimated rotation matrix and translation vector, $\mathbf{R}$ and $\mathbf{t}$ are the ground-truth ones.

Our network is a multi-task network including calculations of output depth map, segmentation mask, 3D-2D correspondences, and 6DoF pose parameters like the current SOTA methods. More generally, when there are multiple target objects in the image, we can estimate the poses of these target objects simultaneously, and the results are given in the experimental section.

\begin{table*}
\centering
\small

\setlength{\tabcolsep}{2mm}
\begin{tabular}{|c|c|c|c|c|c|c|c|c|c|c|}
\hline 2D-3D extractor & PnP type & Ape & Can & Cat  & Driller &Duck & Eggbox(s)  & Glue(s) & Holepun &\bf{Mean}\\
\hline 
\multirow{5}{*}{DGECN\textbf{(Ours)}}
 &DG-PnP(\bf{Ours})  &54.3 &75.9&22.4 &77.5 &51.2 &57.8 &66.9&63.2 &58.7\\
 &PointNet-like PnP\cite{hu2020single}  & 44.4 &71.3&18.5 &71.6 &48.6&51.3 &59.1 &60.3 &53.1\\
&Patch-PnP\cite{Wang_2021_CVPR}  & 51.2 &74.6&21.6 &73.4 &48.5 &56.9 &65.1 &61.4 &56.6\\
&RANSAC-based PnP\cite{lepetit2009epnp}&41.3&66.5&14.3&65.4&44.1&48.9&55.4&56.2&49.0\\
&BPnP\cite{chen2020end}   &46.2  &73.3 & 19.5&72.4&46.2&52.1&61.4&56.2 &53.4 \\
\hline
\multirow{5}{*}{PVNet\cite{peng2019pvnet}}
 &DG-PnP(\bf{Ours})       & 23.4  &68.9 &  23.2  &72.2  &  27.8  &  55.1  &  53.2  &   47.2 &46.4\\
& PointNet-like                     & 19.2&65.1 &  18.9  &  69.0  &  25.3 & 52.0&   51.4&   45.6 &   43.3 \\
&Patch-PnP                          &14.4 & 55.3   & 14.9   &68.2  &   22.1  &   45.9  &  49.4 & 41.3 &   38.9\\
&RANSAC-based PnP        &15.8 & 63.3  &   16.7  &   65.7 &   25.2  &   50.3 &   49.6 &   36.1 & 40.8\\
&BPnP                                   &21.4 & 45.3  &   12.7  &   64.3 &   21.4  &   42.1&  44.5 &   38.7 &36.3 \\
\hline
\multirow{5}{*}{SegDriven\cite{hu2019segmentation}}
 & DG-PnP(\bf{Ours})   & 17.5  & 51.4 &15.9 & 57.9 &20.6  & 31.8&43.2& 39.6&34.7 \\
 &PointNet-like                & 14.8 &45.5 &12.1 &54.6 &18.3   &30.2&45.8 &37.4&32.3\\
 & Patch-PnP                    &  9.8    &36.9  &14.6& 57.3 & 11.6  &28.3&42.3&32.4&28.4  \\
 &RANSAC-based PnP   & 12.1   &39.9& 8.2 &  45.2 &  17.2  &  22.1 &   35.8 &   36.0 &   27.0\\ 
 &BPnP                              & 15.6   &47.8& 14.5 & 51.3 &  14.8  &  30.5 & 26.4 &32.1 &   29.1 \\
 \hline
 \multirow{5}{*}{GDR-Net\cite{Wang_2021_CVPR}}
   &DG-PnP(\bf{Ours})   & 37.5   &78.5&26.8 &  70.6 &  42.9  &  56.8 &   50.4 &   56.4 &   52.5\\
   &PointNet-like PnP      & 17.9   &65.3&18.6 & 62.8  &  31.5  &  48.6 &   36.7 &   49.2 &   41.3 \\
  &Patch-PnP                   &39.3    &79.2& 23.5 &  71.3 & 44.4 &58.2    &  49.3  &  58.7 &   53.0\\
   &RANSAC-based PnP & 20.9   &67.5& 23.9 &  66.1 &  34.9  &  53.4 &   42.3 &   54.3 &   45.4 \\
&BPnP                              & 35.5   &74.2& 21.5 &  67.4 &  36.9  &  51.4 &   45.8 &   51.1 &  48.0 \\
 
 \hline

 \hline  
\end{tabular}
\caption{ 
\textbf{Ablation Study.} Results for different versions of our model with comparison to some baseline models. We evaluate the impact of the DGECN, and DG-PnP. (s) denotes symmetric objects, metrics indicated by light red is the best result. We report the Average Recall (\%) of ADD(-S) on LM-O dataset.
}\label{tab1}
\end{table*}

\section{Experiments}
\label{sec:5}
In this section, we conduct experiments to prove the effectiveness of DGECN. We evaluate our DGECN on several common benchmark datasets. For direct comparison to classic PnP and some learning PnP, we set up several experiments following~\cite{Wang_2021_CVPR,hu2020single} on a synthetic sphere dataset to verify the proposed DG-PnP. Further, we conduct an ablation study to discuss the effectiveness of each component in the proposed method.

\begin{figure*}[t]
\centering
   \includegraphics[width=1.0\linewidth]{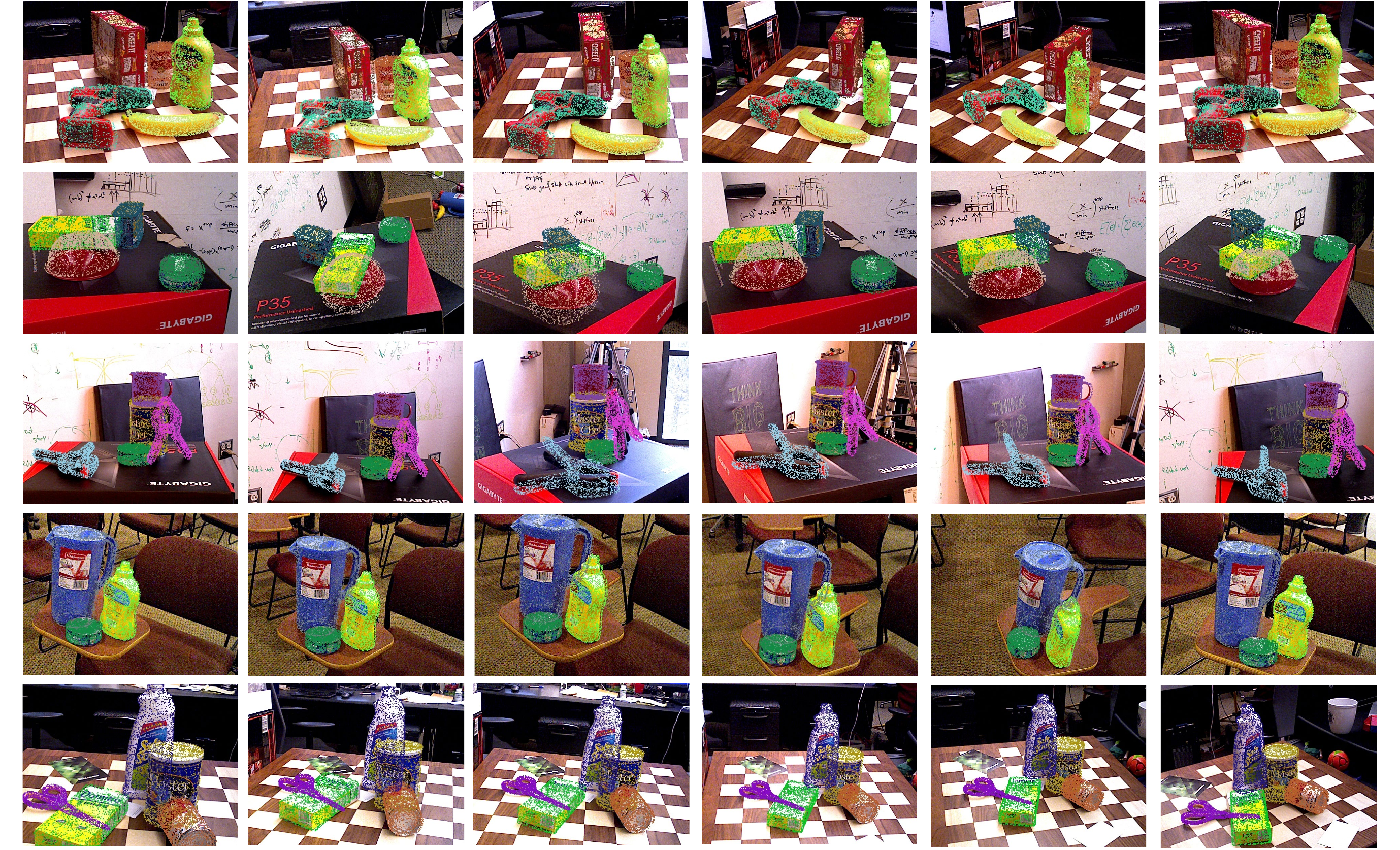}
   \caption{{\bf Qualitative results on YCB-V dataset.} Here we show visualizations of results on YCB-V dataset. Points on different meshes in the same scene are in different colors which projected back to the image after being transformed by the predicted pose.}
\label{fig:7}
\end{figure*}

\subsection{Datasets}
\subsubsection{Synthetic Sphere Dataset.} 
\begin{figure}[t]
\centering
   \includegraphics[width=1.0\linewidth]{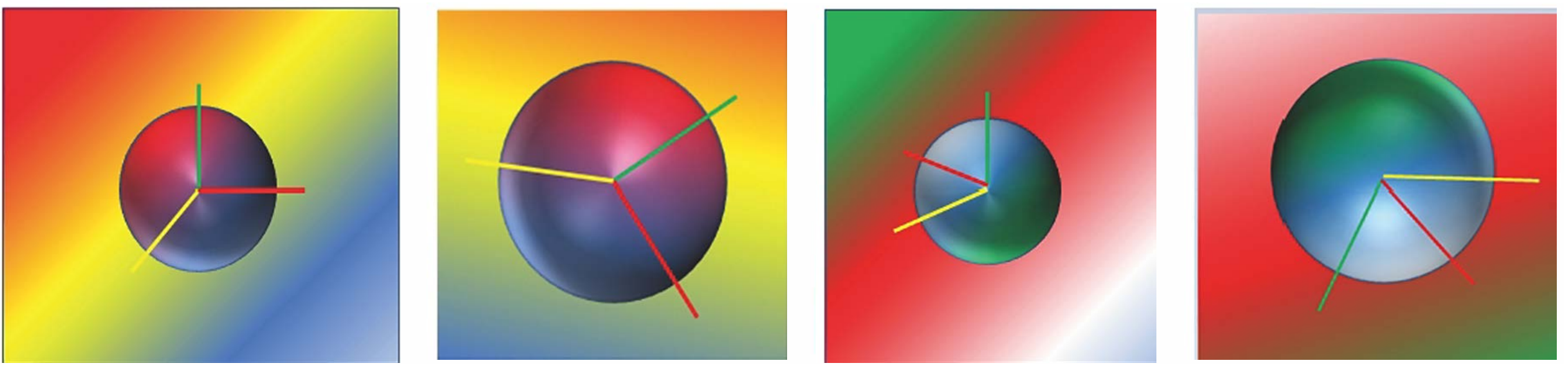}
   \caption{{\bf Synthetic data.} We create synthetic data as in~\cite{hu2020single}, but we add background on theirs. }
\label{fig:4}
\end{figure}

As in Single-Stage~\cite{hu2020single}, we create the exact synthetic 3D-to-2D correspondences using a virtual calibrated camera, with image size of $640\times 480$, focal length of $800$, and principal point at the image center. However, Single-Stage does not require color information, so their background is pure. As discussed in Sec. 3, our network will fully extract local features, including location and color. So we add a gradient background to their synthetic dataset, and the other parameter settings are the same with Single-Stage, as shown in Fig.~\ref{fig:4}.

\begin{figure}[t]
\centering
   \includegraphics[width=1.0\linewidth]{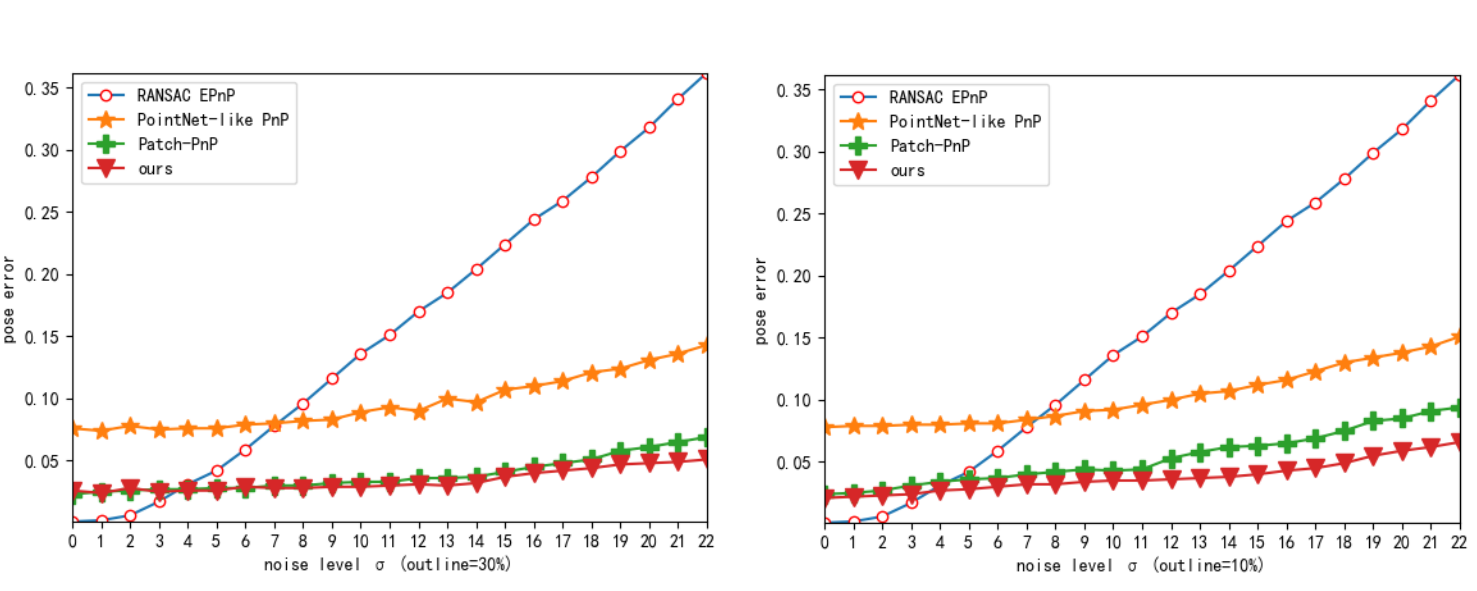}
   \caption{{\bf Comparison with PnP variants }. We compare our method with EPnP~\cite{lepetit2009epnp}, PointNet-like PnP~\cite{hu2020single} and Patch-PnP~\cite{Wang_2021_CVPR}. Our method performs better than PointNet-like PnP all time, and as the noise increases our method is much more accurate and robust than EPnP. The pose error is calculated by ADD.} 
\label{fig:5}
\end{figure}

\subsubsection{YCB-V Dataset.} This dataset is proposed by~\cite{calli2015ycb,xiang2017posecnn} and consists of 21 YCB objects with different shapes and textures. 92 RGB-D videos of the subset of objects were captured and annotated with 6D pose and instance semantic mask. The varying lighting conditions, significant image noise, and occlusions make this dataset challenging. As in PoseCNN~\cite{xiang2017posecnn}, we split the dataset into 80 videos for training and a set of 2,949 keyframes chosen from the rest 12 videos for testing.

\subsubsection{LM-O Dataset.} This dataset~\cite{brachmann2014learning} is a standard benchmark for object 6D pose estimation and contains 13 low-textured objects in 13 videos, annotated 6D pose and instance mask. The main challenges of LM-O are the chaotic scenes, texture-less objects, and lighting variations. In this work, we follow prior works to handle this dataset, and we also add synthesised images into our training set as in~\cite{xiang2017posecnn}.

\subsection{Evaluation metrics}
For comparison, we evaluate our method with two common metrics: the average distance (ADD)~\cite{xiang2017posecnn} and the 2D reprojection error (REP)~\cite{hu2019segmentation}. 

{\bf ADD} uses the average distance between the 3D model points transformed using the predicted pose and those obtained with the ground-truth one. When the distance is less than $10\%$ of the model's diameter, it claims that the estimated pose is correct. We follow~\cite{hu2020single,Wang_2021_CVPR} and evaluate the symmetric object by ADD(-S) metric, which measures the deviation to the closet model point. Denote the predicted pose as $[\mathbf{R^*},\mathbf{t^*}]$ and the ground truth pose as $[\mathbf{R},\mathbf{t}]$:
 \begin{equation}
 \mathrm {ADD}=\frac{1}{m}\sum_{x\in \mathcal{O}}\| (Rx+t)-(R^*x+t^*)\| 
 \end{equation}
 \begin{equation}
 \mathrm {ADD}-\mathrm{S}=\frac{1}{m}\sum_{x_{1}\in \mathcal{O}}\underset{x_{2}\in \mathcal{O}}{min} \| (Rx_{1}+t)-(R^*x_{2}+t^*)\|
 \end{equation}
where $x$ is a vertex of totally $m$ vertices on object mesh $\mathcal{O}$.
When evaluating on YCB-V, we also compute the AUC (area under curve) of ADD(-S) by varying the distance threshold with a maximum of 10 cm~\cite{xiang2017posecnn}.

{\bf REP} computes the mean distance between the projections of 3D model points given the estimated and the ground truth pose. When the REP is below 5 pixels, we claim that the estimated pose is correct.

For each metric, we use the symmetric version for symmetric objects, which we denote by a superscript (s).

\subsection{Comparison with State-of-the-arts}
We compare with the state-of-the-art works on YCB-V and LM-O datasets. It is worth mentioning that we also make a comparison with the RGB-D based methods to verify the effectiveness of our depth estimation network. 

\subsubsection{Performance on LM-O dataset.}  Tab.~\ref{tab2} shows the results of DGECN compared with the state-of-the-art monocular methods on Occlusion LM-O dataset. Our DGECN is comparable to ~\cite{Wang_2021_CVPR,li2018deepim,Di_2021_ICCV} and outperforms ~\cite{hu2020single,peng2019pvnet}. Tab.~\ref{tab5} presents the results of compared with RGB-D based methods. Moreover, in some scenes, the proposed method even outperforms the RGB-D based methods.

\begin{table*}[t]
 \centering
\setlength{\tabcolsep}{1.5mm}{
 \begin{tabular}{cccccccccc}
    \hline
     Method&PoseCNN&PVNet&Single-Stage&HybridPose&GDR-Net&SO-Pose&DeepIM(R)&DPOD(R)&Ours\\
    \hline
Ape        &9.6 &15.8&19.2&20.9&41.3&46.3&\textbf{59.2}&-&50.3\\
Can        &45.2&63.3&65.1&75.3&71.1&\textbf{81.1}&63.5&-&75.9\\
Cat        &0.9 &16.7&18.9&24.9&23.5&18.2&26.2&-&\textbf{26.4}\\
Driller    &41.4&65.7&69.0&70.2&54.6&71.3&55.6&-&\textbf{77.5}\\
Duck       &19.6&25.2&25.3&27.9&41.7&43.9&52.4&-&\textbf{54.2}\\
Eggbox(s)   &22.0&50.2&52.0&52.4&40.2&46.6&\textbf{63.0}&-&57.8\\
Glue(s)     &38.5&49.6&51.4&53.8&59.5&63.3&\textbf{71.7}&-& 66.9\\
Holepun    &22.1&36.1&45.6&54.2&52.6&62.9&52.5&-&\textbf{60.2}\\
    \hline
 Mean     &24.9&40.8&43.3&47.5&47.4&54.3&55.5&47.3&\textbf{58.7} \\
    \hline
  \end{tabular}}
  \caption{Quantitative comparison on known categories of LM-O dataset with state-of-the-art RGB methods with the metric as ADD(-S), (R) stands for Refinement. All methods trained with $real+syn$ data.}
  \label{tab2}
\end{table*}

\begin{table}[t]
 \centering
\setlength{\tabcolsep}{0.6mm}{
 \begin{tabular}{c|c|ccc}
    \hline
     Method&Ref.&ADD(-S)& REP-5px&AUC of ADD-S\\
    \hline
PoseCNN\cite{xiang2017posecnn}&&21.3 &3.7&75.9\\
GDR-Net\cite{Wang_2021_CVPR}&&60.1&-&\textbf{91.6}\\
SO-Pose\cite{Di_2021_ICCV}&&56.8&-&90.9\\
PVNet\cite{peng2019pvnet}&&-&47.4&73.4\\
SegDriven\cite{hu2019segmentation}&&39.0 &30.8&-\\
Single-Stage\cite{hu2020single} &&53.9&48.7&- \\
\hline
DeepIM\cite{li2018deepim} &w/&-&-&88.1 \\
CosyPose\cite{labbe2020cosypose}&w/&-&-&89.8 \\
\hline
Ours&&\textbf{60.6}&\textbf{50.3}&90.9\\
    \hline
  \end{tabular}}
  \caption{Evaluation with state-of-the-art RGB methods on YCB-V. Ref. stands for Refinement.}
  \label{tab3}
\end{table}

\begin{table}[t]
 \centering
\setlength{\tabcolsep}{0.6mm}{
 \begin{tabular}{|c|c|c|c|}
    \hline
     Corr. Extractor&DG-PnP&ADD& AUC of ADD-S\\
    \hline
w/&w/&58.7&90.9\\
w/&w/o&53.2&83.5\\
w/o&w/&50.6&81.3\\
w/o&w/o&41.3&75.3\\
    \hline
  \end{tabular}}
  \caption{\textbf{Ablation on Depth Map.} w/ denotes test with depth map and w/o denotes test without depth map. }
  \label{tab4}
\end{table}

\begin{table}[t]
 \centering
\setlength{\tabcolsep}{0.6mm}{
 \begin{tabular}{cccc}
    \hline
     Method&ADD(-S)& REP-5px&AUC of ADD-S\\
    \hline
Implicit ICP\cite{sundermeyer2018implicit}&64.7 &-&-\\
SSD-6D ICP\cite{kehl2017ssd}&79.0&-&91.6\\
PointFusion\cite{xu2018pointfusion}&-&73.7&73.4\\
DenseFusion\cite{wang2019densefusion}&86.2&30.8&-\\
PVN3D\cite{he2020pvn3d} &53.9&99.4&- \\
\hline
Ours&60.6&50.3&90.9\\
    \hline
  \end{tabular}}
\caption{Evaluation with state-of-the-art RGB-D methods on YCB-V.}
  \label{tab5}
\end{table}

\subsubsection{Performance on YCB-V}
Tab.~\ref{tab3} shows the evaluation results for YCB-V dataset. It shows that our model is comparable to the state-of-the-arts~\cite{labbe2020cosypose,Wang_2021_CVPR} and even outperforms the refinement-based method~\cite{li2018deepim}. Fig.~\ref{fig:7} demonstrates qualitative results on YCB-V.

\subsection{Ablation study}
In this section, we would like to discuss the following questions: (1) How does the DG-PnP compare to the handcrafted PnP and other learnable PnP? (2) Does the learned depth improve the final pose estimation? (3) Is the DGECN effective with PnP variants? 

\textbf{Comparison to PnP variants.} We take 20K synthetic images for training and 2K images for testing. While training, we randomly add 2D noise with variance $\sigma$ in the range of $[0,15]$ and create outliers with $10\%$ and $30\%$. Comparison in synthetic is critical, because it can directly compare our DG-PnP with PnP variants and ignore the influence of the keypoints detection methods. Fig.~\ref{fig:5} shows the results at different noise levels, compared with EPnP~\cite{lepetit2009epnp}, PointNet-like PnP~\cite{hu2020single} and Patch-PnP~\cite{Wang_2021_CVPR}. While handcrafted PnP is more accurate when the noise is minimal, learnable PnP methods are more robust to noise, and they are more accurate when the noise increasing. Moreover, DG-PnP is significantly more robust and accurate than PointNet-like PnP, and comparable with Patch-PnP. Because DG-PnP and Patch-PnP both take into account the geometric and topology features.

\textbf{Ablation on depth map.} As mentioned above, depth information plays a significant role in 6D pose regression. Furthermore, we train our DGECN by discarding the depth estimation. The depth information is used in both correspondence extraction and DG-PnP, so we setup a ablation study on it. As shown in Tab.~\ref{tab4}. DGECN is significantly more robust with depth prediction.

\textbf{Effectiveness of each component.} As shown in tab.~\ref{tab1}, we demonstrate the effectiveness of each component of the proposed method by combining our components with different state-of-the-art methods. For DGECN, we replace the DG-PnP in our architecture with PnP variants~\cite{hu2020single,Wang_2021_CVPR,chen2020end}. DGECN demonstrates a competitive performance with different PnP methods. Moreover, it is even better than Single-Pose combined with the PointNet-like PnP. As for DG-PnP, we replace the PnP variants in some two-stage methods with DG-PnP.

\section{Conclusion}

In this work, we propose a novel depth-guided network for monocular 6D object pose estimation. The core idea is to utilize geometric and topology information, and jointly handles 2D keypoint detection and 6D pose estimation. Then, we delve into 2D-3D correspondences and observe that graph structure can better model the feature of keypoint distributions. Furthermore, we propose a dynamic graph PnP for learning 6D pose to replace the handcrafted PnP. Thus, our approach is a real-time, accurate and robust monocular 6D object pose estimation method.

\section{Acknowledgments}
This work is partially supported by the Key Technological Innovation Projects of Hubei Province (2018AAA062), National Nature Science Foundation of China (NSFC No.61972298) and Wuhan University Huawei GeoInformatics Innovation Lab. 

{\small
\bibliographystyle{ieee_fullname}
\bibliography{egbib}
}

\end{document}